\documentclass[runningheads]{llncs}
\usepackage[T1]{fontenc}
\usepackage{graphicx}
\usepackage{amsmath}
\usepackage{booktabs}

\usepackage[hidelinks]{hyperref}
\usepackage{color}

\usepackage{mathtools}
\begin{document}
\title{Domain-Adapted Latent Diffusion Models for\\ CT-to-PET Image Translation}
\titlerunning{Domain-Adapted Latent Diffusion Models for CT-to-PET Image Translation}
%
%
\author{Yann Bourdé*\inst{1} \and
Lea Bogensperger\inst{1} \and
Arnela Hadzic\inst{2} \and
Michael Krauthammer\inst{1}\\ *yann.bourde@uzh.ch}

\authorrunning{Y. Bourdé et al.}
%

\institute{Department of Quantitative Biomedicine, University of Zurich \and
Institute for Medical Informatics, Statistics and Documentation, Medical University of Graz}

\maketitle              
\begin{abstract}
\sloppy
Positron Emission Tomography (PET) provides important metabolic information for tumor detection and staging but remains significantly less accessible than Computed Tomography (CT) due to its high cost, infrastructure requirements, and reliance on injected radioactive tracers. CT-to-PET image translation using generative models aims to address this disparity, yet state-of-the-art diffusion-based methods are limited by the scarcity of paired training data. 
To overcome this, we propose a domain-adapted latent diffusion framework that transfers natural-image priors using a relatively small dataset of 21,930 paired CT/PET slices, incorporating a signed power transform to resolve extreme intensity sparsity in PET data.
Compared to CPDM, a state-of-the-art baseline for CT-to-PET image translation, our approach yields superior performance in pixel-wise reconstruction and localized tumor fidelity, demonstrating effective transfer of natural-image priors to the medical domain. 
We further provide a systematic analysis of evaluation metric reliability for sparse PET data, showing that standard full-image metrics can be misleading indicators of translated image quality, and propose background-filtered and mask-restricted metrics as more clinically meaningful alternatives.
\end{abstract}
\section{Introduction}

Positron Emission Tomography (PET) and Computed Tomography (CT) are complementary medical imaging modalities that capture anatomical and metabolic information, \textit{e.g.}, for tumor detection, staging, and treatment monitoring. 
Specifically, CT scans provide high-resolution anatomical information through X-ray attenuation, while PET scans reveal metabolic activity via radiotracer distribution, typically $^{18}$F-fluorodeoxyglucose (FDG). Used together, they offer a more complete diagnostic picture than either modality alone. However, PET scanners require costly infrastructure and radiotracer production facilities, making PET imaging less accessible than CT, particularly in low-resource healthcare settings.

With the successful introduction and scalability of generative models~\cite{goodfellow2020generative,kingma2013auto,song2019generative,lipman2022flow} and their numerous applications to the medical imaging domain~\cite{armanious2020medgan,bogensperger2025flowsdf,hadzic2024synthetic}, these methods are also being investigated for paired image-to-image translation tasks such as CT-to-PET image translation~\cite{ref_cpdm}. 
Early approaches relied on Generative Adversarial Networks (GANs)~\cite{ref_isola,ref_zhu}, which suffered from training instability and mode collapse, while modern state-of-the-art (SOTA) approaches rely on diffusion-based models~\cite{ref_cpdm,ref_3dldm}. In particular, latent diffusion models (LDMs)~\cite{ref_ldm} offer a compelling option due to the learned latent space that they operate in, which allows generation of high-resolution images at lower computational cost.
For CT-to-PET translation specifically, Nguyen et al.~\cite{ref_cpdm} recently introduced the CT-to-PET Diffusion Model (CPDM), a conditional diffusion model based on a Brownian Bridge process that defines a stochastic bridge between CT and PET latents rather than corrupting toward pure Gaussian noise, and additionally incorporates domain knowledge through attention and attenuation map modules. Notably, the authors report that LDM priors were inferior to CPDM in their experiments~\cite{ref_cpdm}—a finding we revisit in this work.

A common limitation across these approaches is the scale of publicly available paired CT/PET datasets~\cite{ref_pclt20k}: current SOTA models are trained from scratch on this scarce data~\cite{ref_cpdm,ref_3dldm}, potentially limiting their performance and ability to generalize beyond the training distribution.

Motivated by this gap, we introduce a domain-adapted latent diffusion framework for CT-to-PET image translation in this work. We demonstrate that transfer learning by finetuning a model pretrained on natural image priors effectively mitigates data scarcity and enhances PET synthesis fidelity. Our contributions are summarized as follows:
\begin{itemize}
\item We implement a transfer learning strategy to adapt natural image priors to the PET domain and introduce a signed power transform to reduce the extreme intensity imbalance in PET images.
\item We demonstrate that our framework yields superior performance in pixel-level reconstruction and localized tumor fidelity compared to the state-of-the-art diffusion-based CPDM~\cite{ref_cpdm}, while offering a data-efficient approach for CT-to-PET image translation.
\item We analyze the limitations of standard metrics for sparse PET data and propose background-filtered and mask-restricted metrics as more clinically meaningful alternatives.
\end{itemize}

\section{Method}

\subsection{Conditional CT-to-PET Synthesis}
Applying the diffusion process directly in pixel space is computationally expensive at high resolutions. Latent Diffusion Models (LDMs)~\cite{ref_ldm} address this by compressing images into a low-dimensional latent space via a KL-regularized variational autoencoder (VAE) and running the diffusion process in this space, substantially reducing computational cost while preserving perceptual quality.

In our framework, we utilize this architecture for conditional CT-to-PET synthesis.
Let $z_\mathrm{PET} = \mathcal{E}(x_\mathrm{PET})$ and $z_\mathrm{CT} = \mathcal{E}(x_\mathrm{CT})$ denote the latent representations of the transformed PET and CT images, respectively. During training, the diffusion process adds Gaussian noise to $z_\mathrm{PET}$ over $T$ timesteps to generate $z_\mathrm{PET}^{(t)}$. To enable conditional synthesis, we concatenate the noisy latent $z_\mathrm{PET}^{(t)}$ with the clean CT latent $z_\mathrm{CT}$, providing the U-Net with a combined 8-channel input. 
The U-Net is then trained to learn the conditional reverse diffusion process:
\begin{equation}
z_\mathrm{PET}^{(t-1)} \sim p_\theta\!\left(z_\mathrm{PET}^{(t-1)} \mid z_\mathrm{PET}^{(t)},\, z_\mathrm{CT}\right).
\end{equation}

\subsection{Domain Adaptation and Finetuning Strategy}
The scarcity of paired CT/PET data can hinder the performance of models trained from scratch in this domain. To address this limitation, we utilize transfer learning by initializing our framework with the Stable Diffusion v1.4 backbone pretrained on natural images  from the large-scale LAION dataset~\cite{ref_laion}.
This allows us to effectively leverage the strong structural and perceptual priors learned during pretraining and adapt this backbone to the PET domain using supervised finetuning. 
To accommodate the additional input from the CT latent $z_{\mathrm{CT}}$, we follow Ke et al.~\cite{ref_ke} and expand the first convolutional layer of the UNet by duplicating the pretrained input weights and halving their values, preserving the expected magnitude of activations at initialization.

To validate the framework's suitability for our medical imaging data, we first investigated the reconstruction accuracy of the pretrained VAE on our out-of-distribution medical imaging data. As shown in Table~\ref{tab:vae_metrics} for the PCLT20K dataset, the reconstruction is sufficiently accurate, so we keep the VAE frozen and only focus on finetuning the denoising U-Net.
\begin{table}
\caption{VAE reconstruction metrics.}\label{tab:vae_metrics}
\centering
\begin{tabular}{lllll}
    \hline
    \textbf{Modality} & \textbf{MAE} $\downarrow$ & \textbf{PSNR} $\uparrow$ & \textbf{SSIM} $\uparrow$ & \textbf{LPIPS} $\downarrow$ \\
    \hline
    CT  & 0.0111 & 33.57 & 0.936 & 0.0616 \\
    PET & 0.0019 & 48.07 & 0.982 & 0.0147 \\
    \hline
\end{tabular}
\end{table}

However, natural-image intensity statistics substantially differ from medical modalities such as PET, which exhibit extreme sparsity (see Section \ref{sec:method:dataset}).
Moreover, this sparsity is a central challenge for diffusion-based generative models, as standard noise prediction losses treat all pixels equally regardless of clinical relevance. While recent work has shown that finetuning pretrained models on domain-specific data can yield competitive results with limited paired samples~\cite{ref_wolleb,ref_pinaya}, successful adaptation requires aligning these distributions. Consequently, we apply a signed power transform
\begin{equation}
x'_\mathrm{PET} = \operatorname{sign}(x_\mathrm{PET})\cdot|x_\mathrm{PET}|^{p}
\label{eq:power}
\end{equation}
to the PET images before passing them through the VAE encoder, controlled by $p \in (0, 1]$. It compresses the dynamic range and spreads near-zero values across a broader interval, making the PET distribution more compatible with the natural-image priors of the pretrained model. The inverse transform is applied at inference to recover the original PET intensity scale.

Another key challenge for LDM transfer learning is catastrophic forgetting of pretrained representations. Partial layer freezing~\cite{xie2023difffit} and parameter-efficient adaptation~\cite{ref_lora} can help mitigate this issue. Consequently, we study two finetuning variants: \textit{unfrozen}, in which all U-Net weights are updated, and \textit{frozen}, in which ResNet convolutional blocks, feed-forward layers, and spatial sampling convolutions are kept fixed while attention layers, normalization layers, and the time embedding remain trainable. We compare both variants in Section~\ref{sec:experiments:results} to assess the trade-off between adaptation flexibility and preservation of pretrained priors.

\subsection{Evaluation Metrics}
For conditional synthesis we report standard full-reference metrics: Mean Absolute Error (MAE $\downarrow$), Peak Signal-to-Noise Ratio (PSNR $\uparrow$), Structural Similarity Index (SSIM $\uparrow$), and Learned Perceptual Image Patch Similarity (LPIPS $\downarrow$)~\cite{ref_lpips}. However, Blau and Michaeli~\cite{ref_blau} showed that perceptual quality and pixel-level fidelity are fundamentally at odds, a trade-off directly relevant to comparing generative models across different metric families. For sparse modalities such as PET, where 95--98\% of pixels are near-zero background, full-image metrics are further confounded by background dominance, masking failures in clinically relevant tumor uptake regions. Therefore, we additionally report Masked MAE, Masked SSIM, and Masked LPIPS over the tumor-mask pixels provided by PCLT20K~\cite{ref_pclt20k}. Masked MAE is computed over the tumor-mask pixels directly; because SSIM and LPIPS are patch-based, Masked SSIM and Masked LPIPS are instead computed on a tight bounding box around the mask region to provide sufficient spatial context. 
Mask-restricted metrics quantify fidelity within the tumor region but do not penalize uptake predicted outside it, so we interpret them alongside full-image metrics rather than in isolation.

\section{Experiments}
\subsection{Dataset}
The PCLT20K dataset~\cite{ref_pclt20k}\footnote{Note that we only had access to the PCLT20K dataset, as the paired CT/PET dataset used in CPDM was not publicly available.} contains 21,930 paired axial 2D slices of CT, PET, and binary tumor masks at $512\times512$ resolution, derived from lung tumor scans of 605 patients. CT intensities are normalized to $[-1,1]$ and follow a broad and roughly symmetric distribution. PET images are far more challenging to model: 95--98\% of pixels are near-zero intensity, yielding a normalized mean of $-0.972$. The data splits are predefined with an 80\%/20\% train/test split, from which we construct our validation set of 80 samples.
\label{sec:method:dataset}

\subsection{Training Details}

To evaluate the impact of our proposed strategy, we analyze four model variants: \textit{base} (unfrozen), \textit{base (frozen)}, \textit{power} (unfrozen), and \textit{power (frozen)}. Power model variants are trained on inputs preprocessed with our proposed signed power transform defined in Equation~\ref{eq:power}, using a coefficient $p=0.25$, whereas the base model variants are trained on untransformed inputs. We use L1 loss for all models during training.

All models are finetuned from the Stable Diffusion v1.4 checkpoint, consisting of a frozen variational autoencoder (VAE) ($\approx$83M parameters, spatial compression factor $8\times$) and a denoising U-Net ($\approx$860M parameters). We used AdamW as the optimizer, with a learning rate of $1\times10^{-4}$. Training employed cosine learning rate scheduling and an exponential moving average (EMA) of model weights with a decay of 0.9999. We adopted the DDPM noise schedule~\cite{ref_ddpm} with $T=1000$ steps for both training and sampling at inference.

\subsection{Conditional CT-to-PET Results}
\label{sec:experiments:results}

\begin{figure}[t]
\includegraphics[width=\textwidth]{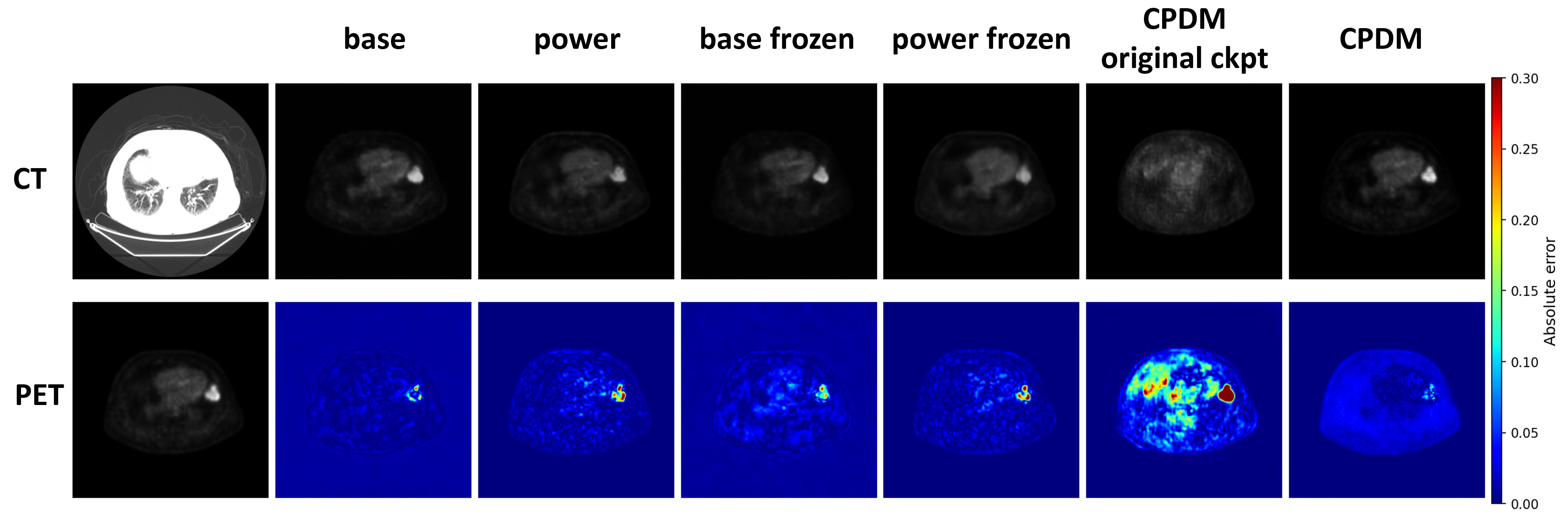}
\caption{Conditional CT-to-PET generation examples. Columns: input CT, then PET generations (top) and absolute error maps vs. ground truth (bottom) for each model.}
\label{fig:cond}
\end{figure}

In our evaluation, we include our four conditional model variants, alongside CPDM's original publicly available checkpoint (\textit{CPDM original ckpt})~\cite{ref_cpdm}, trained on CPDM's unreleased dataset, and a variant of CPDM retrained on the PCLT20K dataset (\textit{CPDM})~\cite{ref_pclt20k}. Table~\ref{tab:cond_full} reports full-image and Table~\ref{tab:cond_mask} reports mask-region metrics for all models evaluated on 416 samples from the PCLT20K test set. Figure~\ref{fig:cond} shows representative conditional generation examples together with pixel-wise error maps.

\newlength{\maecol}\settowidth{\maecol}{0.0204}
\newlength{\psnrcol}\settowidth{\psnrcol}{26.67}
\newlength{\ssimcol}\settowidth{\ssimcol}{0.828}
\newlength{\lpipscol}\settowidth{\lpipscol}{0.145}

\begin{table}
\caption{Conditional CT-to-PET results — full-image metrics, mean and standard deviation. Per-column best results are denoted in \textbf{bold}, second-best results are \underline{underlined}. $^*$ indicates a statistically significant improvement ($p<0.05$, paired Wilcoxon signed-rank test) compared to the retrained CPDM baseline.}
\label{tab:cond_full}
{\fontsize{8}{9}\selectfont
\begin{tabular*}{\textwidth}{@{\extracolsep{\fill}}lllll}
\hline
\textbf{Model} & \multicolumn{1}{c}{\textbf{MAE}$\downarrow$} & \multicolumn{1}{c}{\textbf{PSNR}$\uparrow$} & \multicolumn{1}{c}{\textbf{SSIM}$\uparrow$} & \multicolumn{1}{c}{\textbf{LPIPS}$\downarrow$}\\
\hline
CPDM original ckpt      & $\mathmakebox[\maecol][r]{0.0204} \pm 0.0079$ & $\mathmakebox[\psnrcol][r]{26.67} \pm 2.80$ & $\mathmakebox[\ssimcol][r]{0.828} \pm 0.060$ & $\mathmakebox[\lpipscol][r]{0.145} \pm 0.020$ \\
CPDM  & $\mathmakebox[\maecol][r]{\underline{0.0141}} \pm 0.0101$ & $\mathmakebox[\psnrcol][r]{29.04} \pm 4.30$ & $\mathmakebox[\ssimcol][r]{0.884} \pm 0.082$ & $\mathmakebox[\lpipscol][r]{\textbf{0.109}} \pm 0.021$ \\
\hline
base         & $\mathmakebox[\maecol][r]{0.0182} \pm 0.0100$ & $\mathmakebox[\psnrcol][r]{\textbf{30.62}} \pm 4.64$ & $\mathmakebox[\ssimcol][r]{0.591} \pm 0.079$ & $\mathmakebox[\lpipscol][r]{0.204} \pm 0.023$ \\
power        & $\mathmakebox[\maecol][r]{\textbf{0.0129}} \pm 0.0081^*$ & $\mathmakebox[\psnrcol][r]{30.04} \pm 4.01$ & $\mathmakebox[\ssimcol][r]{\textbf{0.913}} \pm 0.066^*$ & $\mathmakebox[\lpipscol][r]{\underline{0.123}} \pm 0.024^*$ \\
base frozen  & $\mathmakebox[\maecol][r]{0.0173} \pm 0.0080^*$ & $\mathmakebox[\psnrcol][r]{\underline{30.17}} \pm 3.89$ & $\mathmakebox[\ssimcol][r]{0.661} \pm 0.069^*$ & $\mathmakebox[\lpipscol][r]{0.238} \pm 0.021^*$ \\
power frozen & $\mathmakebox[\maecol][r]{0.0158} \pm 0.0072$ & $\mathmakebox[\psnrcol][r]{28.50} \pm 3.22$ & $\mathmakebox[\ssimcol][r]{\underline{0.893}} \pm 0.062$ & $\mathmakebox[\lpipscol][r]{0.137} \pm 0.023$ \\
\hline
\end{tabular*}}
\end{table}

\newlength{\mmaecol}\settowidth{\mmaecol}{0.189}
\newlength{\mssimcol}\settowidth{\mssimcol}{0.243}
\newlength{\mlpipscol}\settowidth{\mlpipscol}{0.463}

\begin{table}
\caption{Conditional CT-to-PET results — masked/tumour-box metrics, mean and standard deviation. Per-column best results are denoted in \textbf{bold}, second-best results are \underline{underlined}. $^*$ indicates a statistically significant improvement ($p<0.05$, paired Wilcoxon signed-rank test) compared to the retrained CPDM baseline.}
\label{tab:cond_mask}
{\fontsize{8}{9}\selectfont
\begin{tabular*}{\textwidth}{@{\extracolsep{\fill}}llll}
\hline
\textbf{Model} & \multicolumn{1}{c}{\textbf{Masked MAE}$\downarrow$} & \multicolumn{1}{c}{\textbf{Masked SSIM}$\uparrow$} & \multicolumn{1}{c}{\textbf{Masked LPIPS}$\downarrow$}\\
\hline
CPDM original ckpt      & $\mathmakebox[\mmaecol][r]{0.189} \pm 0.120$ & $\mathmakebox[\mssimcol][r]{0.243} \pm 0.124$ & $\mathmakebox[\mlpipscol][r]{0.463} \pm 0.066$ \\
CPDM  & $\mathmakebox[\mmaecol][r]{0.221} \pm 0.131$ & $\mathmakebox[\mssimcol][r]{0.276} \pm 0.160$ & $\mathmakebox[\mlpipscol][r]{0.400} \pm 0.069$ \\
\hline
base         & $\mathmakebox[\mmaecol][r]{0.181} \pm 0.101$ & $\mathmakebox[\mssimcol][r]{\textbf{0.485}} \pm 0.194$ & $\mathmakebox[\mlpipscol][r]{\textbf{0.368}} \pm 0.098$ \\
power        & $\mathmakebox[\mmaecol][r]{\textbf{0.165}} \pm 0.092^*$ & $\mathmakebox[\mssimcol][r]{\underline{0.480}} \pm 0.175^*$ & $\mathmakebox[\mlpipscol][r]{\underline{0.369}} \pm 0.092^*$ \\
base frozen  & $\mathmakebox[\mmaecol][r]{\underline{0.171}} \pm 0.098^*$ & $\mathmakebox[\mssimcol][r]{0.462} \pm 0.170^*$ & $\mathmakebox[\mlpipscol][r]{0.413} \pm 0.094^*$ \\
power frozen & $\mathmakebox[\mmaecol][r]{\textbf{0.165}} \pm 0.095$ & $\mathmakebox[\mssimcol][r]{0.427} \pm 0.157$ & $\mathmakebox[\mlpipscol][r]{0.416} \pm 0.090$ \\
\hline
\end{tabular*}}
\end{table}

The power transform consistently improves full-image MAE, SSIM, and LPIPS, indicating that addressing PET sparsity at the preprocessing stage is beneficial. Freezing a subset of U-Net layers has negligible effect on performance, suggesting that ResNet convolutional and feed-forward features transfer effectively from natural image pretraining and that adaptation is largely carried by the trainable attention layers. 
Our best model (\textit{power}) outperforms both CPDM checkpoints on MAE and SSIM, and PSNR. CPDM achieves better LPIPS, which likely reflects the perception-distortion tradeoff~\cite{ref_blau}. CPDM's Brownian Bridge formulation may favor perceptual realism over pixel-level fidelity, while our model optimizes more directly for reconstruction accuracy. In the context of medical imaging, pixel-level fidelity may be preferable for preserving clinically relevant intensity information, though this would require clinical validation.

Mask-region metrics are substantially lower than full-image metrics for all models, highlighting the difficulty of accurately localizing and synthesizing PET uptake in tumor regions. Our models perform very similarly to each other across the three masked metrics. Notably, the large SSIM gap between \textit{base} and \textit{power} seen on full-image metrics largely disappears here. This suggests that the power transform's global gains may come at a slight cost within the highest-intensity tumor regions.

Overall, our results validate our transfer learning framework as a robust approach for CT-to-PET synthesis, demonstrating that targeted intensity transformations and selective finetuning can successfully adapt pretrained generative models to this cross-modality task.

\subsection{Metric Caveats and Background Filtering}

A key finding of this work is that standard full-image metrics are unreliable for sparse PET evaluation. Because 95--98\% of PET pixels are near-zero background, a model perfectly reproducing background while misplacing tumor uptake can still achieve low MAE and high SSIM, while a clinically superior model that correctly localizes uptake but introduces minor background variation appears worse.

\begin{table}
\caption{Conditional CT-to-PET evaluation metrics with background filter ($\tau = 0.05$), averaged over 160 test samples.}\label{tab:filter_metrics}
\centering
{\fontsize{8}{9}\selectfont
\begin{tabular}{llllll}
\hline
 & \textbf{MAE $\downarrow$} & \textbf{PSNR $\uparrow$} & \textbf{SSIM $\uparrow$} & \textbf{LPIPS $\downarrow$} & \textbf{Masked MAE $\downarrow$} \\
\hline
Unfiltered (mean) & 0.0154 & 32.58 & 0.587 & 0.184 & 0.1719 \\
Filtered (mean)   & 0.0093 & 32.91 & 0.909 & 0.152 & --- \\
\hline
\end{tabular}}
\end{table}

\begin{figure}[t]
\includegraphics[width=\textwidth]{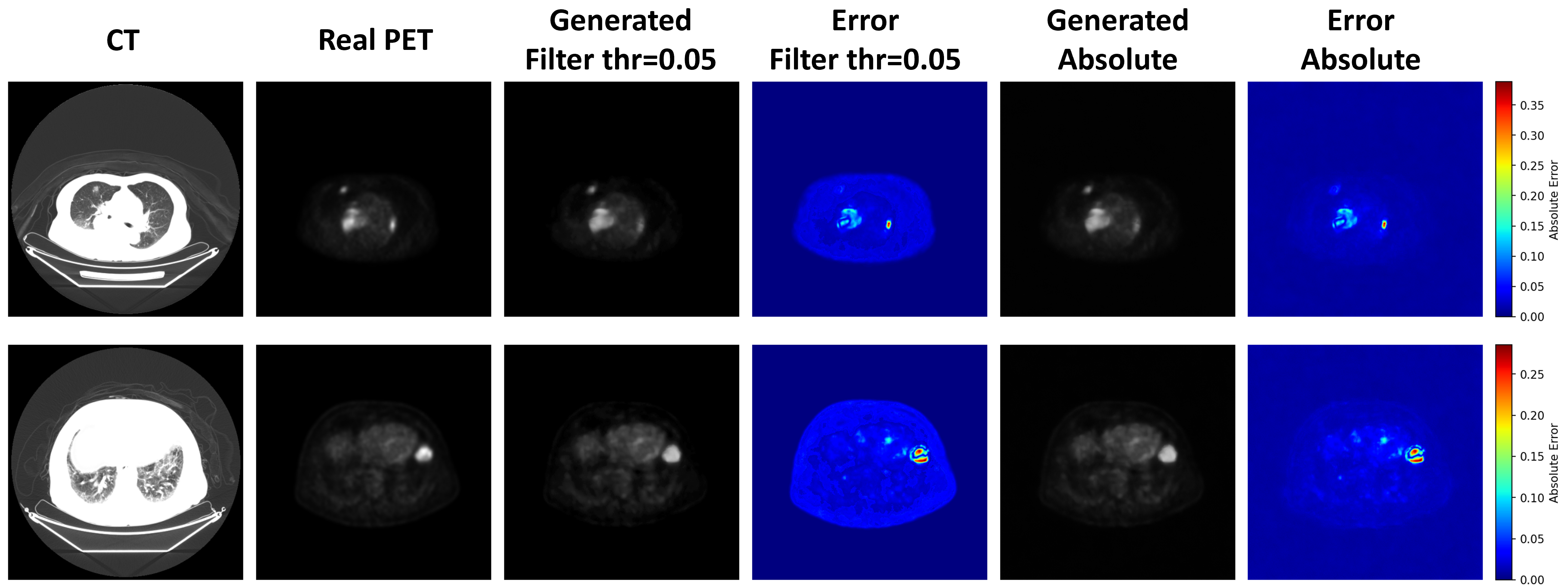}
\caption{Filtered vs unfiltered CT-to-PET generations with absolute error maps.}
\label{fig:filter}
\end{figure}

To quantify this effect, we evaluated our \textit{base} conditional model under a background filter that zeroes all pixels below an absolute intensity threshold of $\tau=0.05$ before computing metrics. Figure~\ref{fig:filter} shows example outputs under this filter and Table~\ref{tab:filter_metrics} reports full-image metrics averaged over 160 test samples, with and without the filter. For reference, the table also includes Masked MAE (0.1719), which is already foreground-restricted, so the background filter does not apply (---). Most metrics significantly improve with the filter, in particular SSIM, because the base model tends to produce light noise in the background which significantly skews full-image metrics. When this noise is filtered we see that the full-image metrics match those of the power models. 
This explains the apparent gap between \textit{base} and \textit{power} in Section~\ref{sec:experiments:results}: it largely reflects background noise rather than a difference in tumor-region fidelity.
These results indicate that the model captures signal-bearing regions well and that degradation in full-image metrics is largely driven by minor background variation rather than failure in clinically relevant regions. In light of these findings, we recommend that future work on PET synthesis adopt background-filtered or mask-restricted metrics as the primary evaluation criterion alongside full-image metrics.

\subsection{Ablations and Extended Experiments}

\subsubsection{Power transform.}\
We evaluated frozen and unfrozen variants using different power coefficients $p=0.125$ and $p=0.5$ and compared them against our $p=0.25$ model variants. The resulting metrics are remarkably similar across all four: SSIM approximately 0.88, PSNR approximately 29, MAE approximately 0.015, and LPIPS between 0.12 and 0.15, with the masked metrics following the same pattern. Overall, our best $p=0.25$ model variant (\textit{power}) remains the top performer with the strongest mean metrics, though the other coefficient variants often fall within its standard deviation range. This result suggests that the optimal coefficient may vary across datasets, and that what matters most is applying the transform at all, rather than the precise value of $p$.

\subsubsection{Statistical significance.}\
We evaluated the statistical significance of our results in Tables~\ref{tab:cond_full} and~\ref{tab:cond_mask} using a paired Wilcoxon signed-rank test across three models: \textit{base frozen}, \textit{power}, and retrained \textit{CPDM}, on our 416 test samples. We tested all full-image and masked metrics across all three pairs and found that the only non-significant result was full-image PSNR between our \textit{base frozen} and \textit{power} models. This confirms the statistical significance ($p < 0.05$) of our results against CPDM, as well as between our own model variants across all other metrics.

\subsubsection{Out-of-distribution experiment.}\
To evaluate robustness to corrupted input, we fed CT samples corrupted with increasing levels of Gaussian noise (0, 0.05, 0.1, 0.25, 0.5, 0.75, 1.0) into our \textit{base frozen} and \textit{power} models and measured the resulting metrics on 16 samples per level. The metrics degrade unevenly across noise levels. LPIPS worsens consistently, from 0.230 to 0.240 for \textit{base frozen} and from 0.143 to 0.187 for \textit{power}. MAE and SSIM instead dip at low-to-moderate noise (0.05, 0.1, 0.25) before partially recovering at higher levels (0.75, 1.0). PSNR shows no consistent trend in either direction, with high standard deviation throughout. Masked metrics degrade more consistently with noise, particularly at the highest levels (0.75 and 1.0). We attribute these differences to what each metric is sensitive to. Since LPIPS measures perceptual realism, as conditioning degrades and the model falls back toward a generic output, the result drifts steadily further from a realistic PET image. MAE and SSIM instead measure direct structural and pixel-wise alignment: at moderate noise levels the model still follows the corrupted conditioning and generates specific, confidently wrong detail, producing large structural error, whereas at high noise the conditioning carries essentially no signal and the model regresses toward a generic output that is structurally closer to typical ground truth on average. PSNR's volatility likely reflects this same tension between detail-following error and regression to the mean, compounded by the small sample size ($n=16$) per level. Masked metrics degrade more steadily because the tumor region depends on fine, localized detail that is lost outright at higher noise levels.

\section{Conclusion}
In this work, we presented a data-efficient transfer learning framework for CT-to-PET image translation, demonstrating that natural-image priors can be effectively adapted using a small dataset of 21,930 paired slices. By incorporating a signed power transform to address inherent PET intensity imbalance, our approach yields superior reconstruction accuracy and localized tumor synthesis when compared to the state-of-the-art CPDM baseline.
We further identified background-pixel dominance as a fundamental limitation of standard evaluation metrics for sparse PET data, and showed that background-filtered and tumor-mask-restricted metrics provide more clinically meaningful assessments. 
These findings suggest that finetuning pretrained latent diffusion models is a data-efficient approach to medical image synthesis, and highlight the need for carefully designed evaluation protocols for sparse medical modalities.

Several directions remain for future work. First, systematically reducing the training set would characterize the minimum number of paired samples required before transfer learning degrades, quantifying the data-efficiency limit of this approach. Moreover, adding losses computed in image space by backpropagating through the VAE decoder---rather than in latent space---may better align training with pixel-level clinical fidelity and could clarify the interaction between loss function choice and the signed power transform observed in this work.

\bibliographystyle{splncs04}
\bibliography{mybibliography}

\end{document}